\documentclass{article}

\usepackage[nonatbib, final]{nips_2017}

\usepackage[utf8]{inputenc} 
\usepackage[T1]{fontenc}    
\usepackage{hyperref}       
\usepackage{url}            
\usepackage{booktabs}       
\usepackage{amsfonts}       
\usepackage{nicefrac}       
\usepackage{microtype}      
\usepackage{enumitem}
\usepackage{hyperref}
\usepackage{euscript}
\usepackage{subcaption}
\usepackage{float}

\usepackage{biblatex} 
\usepackage{graphicx}

\title{Advances in Experience Replay}

\author{
  Hima Tammineedi\\
  Department of Computer Science\\
  Carnegie Mellon University\\
  Pittsburgh, PA 15213 \\
  \texttt{htammine@andrew.cmu.edu}
  \And
  Tracy Wan\\
  Department of Computer Science\\
  Carnegie Mellon University\\
  Pittsburgh, PA 15213 \\
  \texttt{twan@andrew.cmu.edu}
   \And
  Neil Xu\\
  Department of Computer Science\\
  Carnegie Mellon University\\
  Pittsburgh, PA 15213 \\
  \texttt{ziyux@andrew.cmu.edu} \\
} 

\begin{document}

\maketitle
\begin{abstract}
This project combines recent advances in experience replay techniques, namely, Combined Experience Replay (CER), Prioritized Experience Replay (PER), and Hindsight Experience Replay (HER). We show the results of combinations of these techniques with DDPG and DQN methods. CER always adds the most recent experience to the batch. PER chooses which experiences should be replayed based on how beneficial they will be towards learning. HER learns from failure by substituting the desired goal with the achieved goal and recomputing the reward function. The effectiveness of combinations of these experience replay techniques is tested in a variety of OpenAI gym environments.
\end{abstract}

\section{Introduction}
With the rapid introduction of new techniques to reinforcement learning, a smörgåsbord of approaches have emerged, all promising improvements over baseline methods. The goal of our project is to examine some of the most high impact recent advances in reinforcement learning, and see how and whether they can be combined to create a new state of the art standard in performance on OpenAI Gym tasks. Our hope is that we can identify which of these methods, or combination of these methods, has the best performance, and create a new benchmark for these standardized learning environments. We will examine techniques involving two popular and promising ideas in current in reinforcement learning literature: experience replay (ER).

With the advent of the successful utilization of deep neural networks (DNNs) as function approximators in various model-free techniques based upon TD learning, experience replay has become a necessary tool to enhance  accurate and generalized learning by DNNs. As a result of the emphasis that has been placed on experience replay, a variety of modifications have emerged in recent years that have individually shown significant increases in convergence speed when applied to DNN-based learning models. The ones that seem to have shown the most dramatic performance improvements are prioritized experience replay (PER) \cite{prioritized} and hindsight experience replay (HER) \cite{hindsight}. Another technique, combined experience replay (CER) \cite{combined} has also been shown to improve performance.

To our knowledge, there has been no study on the combination of all the experience replay advances in recent years. Thus, We combine the recent experience replay techniques of HER, PER, and CER in order to show the combined effectiveness in a multitude of environments.

\section{Background and Related Works}

\subsection{Deep Q Network}
In standard reinforcement set up, an agent interacts with an environment in discrete time steps. At each time t, the agent receives an observation $x_t,$ makes an action $a_t$ and receives a reward $r_t.$ An agent's behavior is defined by a policy $\pi$ which given a state outputs a probability distribution over the possible actions. The process is modeled as a Markov Decision process with state space $\EuScript{S}$ action space $\EuScript{A}$ and an initial state distribution $p(s_0),$ transition dynamics $p(s_{t+1}|s_t,a_t),$ and reward function $r(s_t, a_t).$ The return is defined as sum of discounted reward $\gamma \in [0,1].$ 
\begin{equation}
R_T = \sigma_{i=t}^{T} \gamma(i-t)r(s_i, a_i) 
\end{equation}
The goal of reinforcement learning is to learn a policy distribution that maximizes the expected reward. The expected reward after taking an action $a_t$ in state $s_t$ following policy $\pi$ is 
\begin{equation}
Q^{\pi} (s_t, a_t) = E_{r_i \geq t, s_{i>t}~E, a_{i>t}~\pi}[R_t|s_t, a_t]
\end{equation}
Expanding the expectation gives the Bellman Equation
\begin{equation}
Q^{\pi}(s_t, a_t) = E_{r_t, s+1 ~ E} [r(s_t, a_t) + \gamma E_{a_{t+1}~\pi}[Q^{\pi}(s_{t+1}, a_{t+1}]]
\end{equation}
If the target policy is deterministic, it can be described as a function $\mu: S -> A$
\begin{equation}
Q^{\mu}(s_t, a_t) = E_{r_t, s+1 ~ E} [r(s_t, a_t) + \gamma Q^{\mu}(s_{t+1}, \mu(s_{t+1}))]
\end{equation}
The expectation depends only on the environment. It is possible to learn $Q^\mu$ off policy, using transitions generated from a different policy. 

Q-learning uses the greedy policy $\mu(s) = argmax_a Q(s,a).$ This can be approximated by minimizing the loss of its parametrization using $\theta^{Q}$ 
\begin{equation}
L(\theta^{Q}) = E_{s_t~p^{\pi}, a_t ~ \pi, r_t~E} [(Q(s_t, a_t | \theta^Q) - y_t)^2]
\end{equation}
where 
\begin{equation}
y_t = r(s_t, a_t) + \gamma Q(s_{t+1}, \mu Q(s_{t+1}, \mu(s_{t+1})|\theta^Q)
\end{equation}
Using replay buffer and a separate target network for calculating t, large neural networks could be used to approximate the Q function. This is known as deep Q learning. 
\subsection{Deep Deterministic Policy Gradients (DDPG)}
The DDPG algorithm \cite{ddpg} uses an actor-critic approach based on the DPG algorithm. The DPG algorithm uses a $\mu(s | \theta^{\mu})$ to specify the deterministic policy by returning an action given a state. The critic $Q(s|a)$ is updated using the Bellman equation. The actor is updated using the following gradient
\begin{equation}
   \Delta_{\theta \mu} J \approx E_{st~p^{\beta}}[\Delta Q(s,a|\theta^{Q})|_{s=s_t,a=\mu(s_t)}\Delta_{\theta_{\mu}(s|\theta^{\mu})|_{s=s_t}}]
\end{equation}
Rather than directly copying the weights, the DDPG algorithm creates a copy of the actor and critic network $Q'(s, a|\theta^{Q'})$ and $\mu'(s|\theta^(\theta'))$ and perform target updates. The weights of the networks are updated by having them slowly match the learned network: $\theta' <- \tau\theta + (1 - \tau)\theta'$ with $\tau << 1.$ The target network is constrained to train slowly, making the entire network more stable. In order for the hyper parameters to generalize across environments with different scales of state value, DDPG employs batch normalization that normalizes each dimension across the samples in a unit to have unit mean and variance. It maintain a running average of the mean and variance to use for normalization. Batch normalization is applied on the state input, all layers of the $Q$ and $\mu$ network prior to the action output. It allows the system to learn on different environment with different settings. The exploration policy samples from a noise process in addition to the actor policy.
\begin{equation}
\mu'(s_t) = \mu(s_t|\theta_t^{\mu}) + \EuScript{N}
\end{equation} 
The noise process is chosen to suit the environment. 

\subsection{Hindsight Experience Replay (HER)}

In HER\cite{hindsight}, the trained value function takes in not only state $s \in S$ but also a goal $g \in G.$ After experiencing some episode $s_0, s_1, ... s_t,$ each transition is stored in the memory replay buffer with both the original goal and the same transition with the original replaced with an alternative goal. Thus, HER is motivated on the principle that an agent that performs multi-task learning (in this case on all the goals in the goal-space $G$) will learn more quickly, then an agent solely trying to learn the singular, original goal. 

In this case, the different set of goals is $m(s_t),$ the goal achieved in the final state of the episode. This is especially useful a sparse reward environment, where an agent training to achieve single goal has trouble receiving any useful reward. By setting the goal to the final state of the episode (or more generally, by setting the goal s.t. the agent achieves a big reward through the episode) the agent will receive much more reward signal in its experiences, and thus learn more quickly.

\subsection{Prioritized Experience Replay (PER)}
In regular experience replay, all transitions are sampled uniformly. But this does not seem ideal if some transitions do not really help the agent learn, yet we keep sampling them.\\
In PER \cite{prioritized}, the transitions are sampled according to how helpful they will be for learning. Clearly, we have no way, at present, of knowing exactly how much each transition will help the network in its learning progress, but we can try to get a proxy for it.\\
Transitions with high expected learning progress, as measured by the magnitude of their temporal-difference (TD) error ($\delta$) are replayed more frequently. The magnitude of the TD error indicates how "surprising" a given transition is since our network did not predict the Q-values well, and so we prioritize these for learning. \\

However, we don't want to only choose the transitions that have the highest priorities as this can lead to a loss of diversity and thus over-fitting. So we ensure that there is a non-zero sampling probability for all transitions (equation \ref{eq:PER}).

The probability of sampling transition $i$ is 
\begin{equation}
    \label{eq:PER}
    P(i) = \frac{p^{\alpha}_i}{\sum_k p^\alpha_k}
\end{equation}
where $p^i$ is the priority of transition $i$ (in this case $p_i = |\delta_i|$).\\

We further use importance sampling weights to correct for the bias introduced by the prioritization changing the original data distribution.

\subsection{Combined Experience Replay (CER)} 
CER \cite{combined} is a special case of PER. PER gives the latest transition a higher priority but it is not guaranteed to be replayed immediately. CER deals with this by adding the latest transition into the training batch. As a hyper parameter, the size of the replay buffer is extremely sensitive to the stabilization of the training system. CER attempts to remedy the effect of having a large replay buffer by ensuring that the latest transition is sampled. 

\subsection{Evaluating experience replay techniques}

Previous work has addressed the issue of memory replay size, by either measuring the empirical results of changing the buffer size on Gym and Atari environments \cite{combined}, or using analytical techniques to derive a theoretically optimal buffer size dynamically throughout training\cite{adaptive}. These works provide insight into the the hyperparameter of the buffer size of the experience, and provide strategies through which that hyperparameter may be chosen to maximize the convergence speed. Zhang et al. (2017) demonstrates that the choice of buffer size can have a large impact on the sample efficiency of the model being trained, and furthermore, proposes that experience replay can be detrimental if used with improper priority methods since it may delay certain samples that could speed up convergence due to the stochastic nature of the sampling process.


\section{Methodology}
We first implemented the three individual experience replay techniques to establish a baselines for how well they can perform. We then tried the combinations of the various techniques as well. To establish the efficacy of these techniques, we first tested these in conjunction with a deep Q-network (DQN) on the CartPole, MountainCar, and LunarLander environments from OpenAI Gym. 
For the DQN, we implemented a DQN with target fixing in order to increase stability. 

We then extended our methods to use continuous environments. Since DQN cannot generate continuous outputs, we implemented a Deep Deterministic Policy Gradient (DDPG).  
We tested DDPG on the Pendulum, Continuous Lunar Lander and Continuous Mountain Car environments. 

We looked at two metrics when evaluating all of our various model depending on the environment: (1) highest reward after a fixed number of episodes or (2) speed of convergence i.e. how many episodes until convergence. where We define convergence to be the first episode where the frozen policy network can achieve an average reward over 100 episodes that exceeds or is equal to the goal for solving the environment. This is defined to be the following for the environments we utilized:

\begin{table}[h]
\centering
\label{solve-reward}
\begin{tabular}{ll}
\textbf{Environment} & \textbf{Reward needed to solve} \\
CartPole-v0 & 200 \\
MountainCar-v0       & -110                            \\
LunarLander-v2       & 200                             \\
\end{tabular}
\end{table}

For the remaining environments i.e. Acrobot-v1 and Pendulum-v0, we measure the effectiveness of our experience replay strategies by setting an episode limit for each environment and analyzing their performance by taking the best average reward over 100 episodes model from the training rule.

For each environment, we tried every possible combination of combined, prioritized, and hindsight experience replay strategies, which in total resulted in 8 different agents being experimented on at most. We wanted to test, even for simple environments, whether the combination of different experience replay techniques could be counterproductive when utilized simultaneously, or that the techniques could all yield improvements in convergence rate and sample efficiency. However, some tasks are not conducive to the goal based formulation that hindsight experience replay uses. Thus, for CartPole-v0, Acrobot-v1, and Pendulum-v0, we only run variants of combined and experience replay, which totals of 4 different agents being run in those environments



All of our code is available at: \url{https://github.com/himat/CHAPtER}

\section{Results}

Below are our results for the various environments we tested our experience replay strategies on. Note that we tested every combination of combined (C), prioritized (P), and hindsight (H) experience replays (ER) for each environment, providing an exhaustive search for the interactions between different types of strategies.

\begin{table}[H]
\centering
\caption{DQN used to solve these discrete action environments (CartPole does not support hindsight goals)}
\label{dqn-results}
\begin{tabular}{l|l|l|l|}
\cline{2-4}
                                                                    & \multicolumn{3}{c|}{\textbf{Environments}}                               \\ \hline
\multicolumn{1}{|l|}{\textbf{Strategies (Episodes to Convergence)}} & \textit{CartPole-v0} & \textit{MountainCar-v0} & \textit{LunarLander-v2} \\ \hline
\multicolumn{1}{|l|}{\textit{Baseline}}                             & 4000                    & 33000                   & 3500                    \\ \hline
\multicolumn{1}{|l|}{\textit{CER}}                                  & 3500                    & 15000                   & 4500                    \\ \hline
\multicolumn{1}{|l|}{\textit{PER}}                                  & 4000                    & N/A                      & 7000                    \\ \hline
\multicolumn{1}{|l|}{\textit{HER}}                                  & N/A                    & 15500                   & 11000                   \\ \hline
\multicolumn{1}{|l|}{\textit{CPER}}                                 &5000                   & N/A                       & 19500                   \\ \hline
\multicolumn{1}{|l|}{\textit{HPER}}                                 & N/A                    & N/A                      & 10000                   \\ \hline
\multicolumn{1}{|l|}{\textit{CHER}}                                 & N/A                    & 17500                   & 4500                    \\ \hline
\multicolumn{1}{|l|}{\textit{CHPER}}                                & N/A                    & N/A                       & 8000                    \\ \hline
\end{tabular}
\end{table}
Interestingly, different environments have different strategies that seem to be more optimal than others. Noticeably for LunarLander-v2, the baseline actually performs the best out of all the experience replay strategies. In fact, CPER, HER, and HPER are perform significantly worse than the baseline.

\begin{figure}[H]
\begin{subfigure}[b]{0.45\textwidth}
\includegraphics[width=\textwidth]{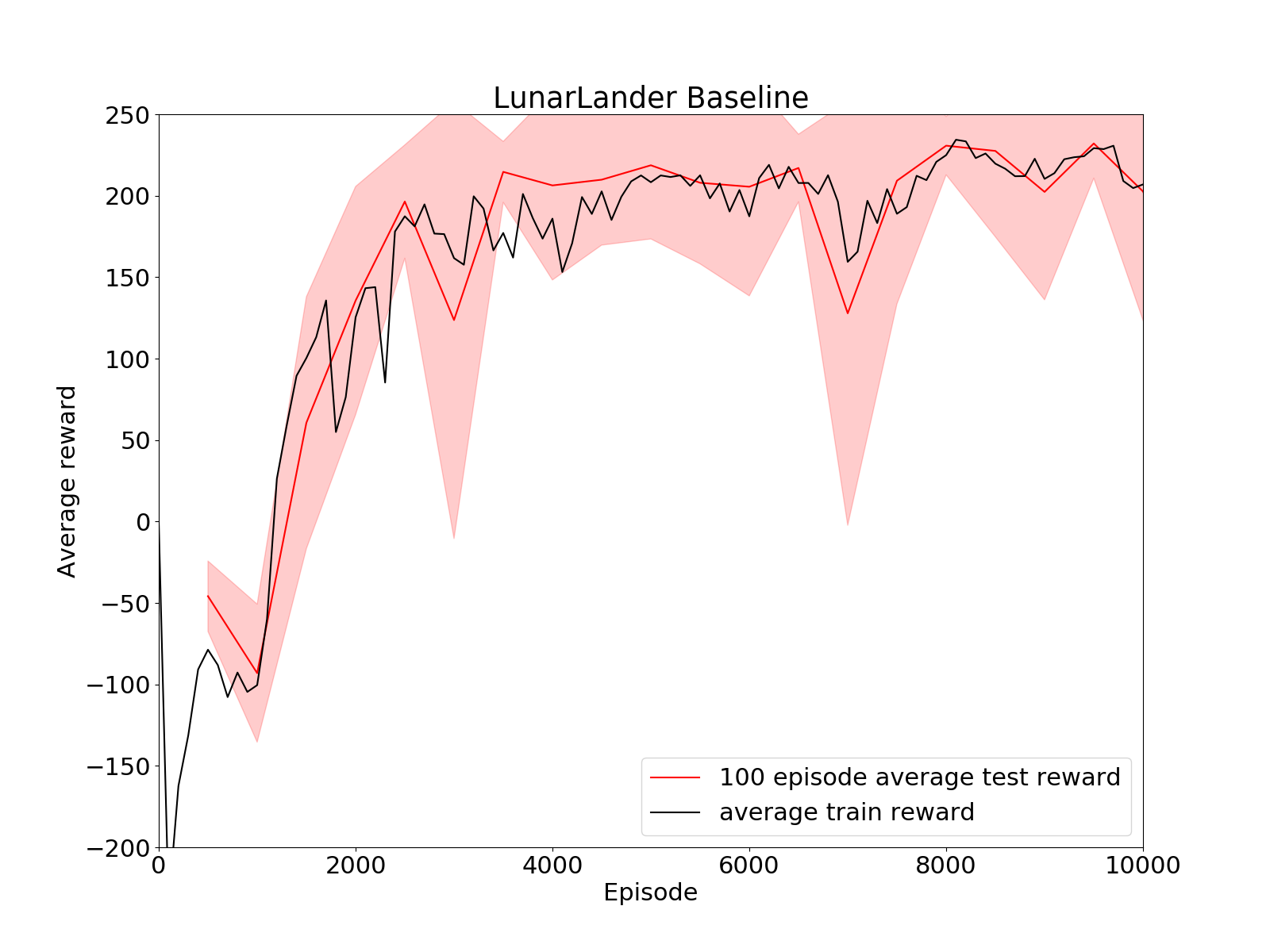}
\end{subfigure}%
\begin{subfigure}[b]{0.45\textwidth}
\includegraphics[width=\textwidth]{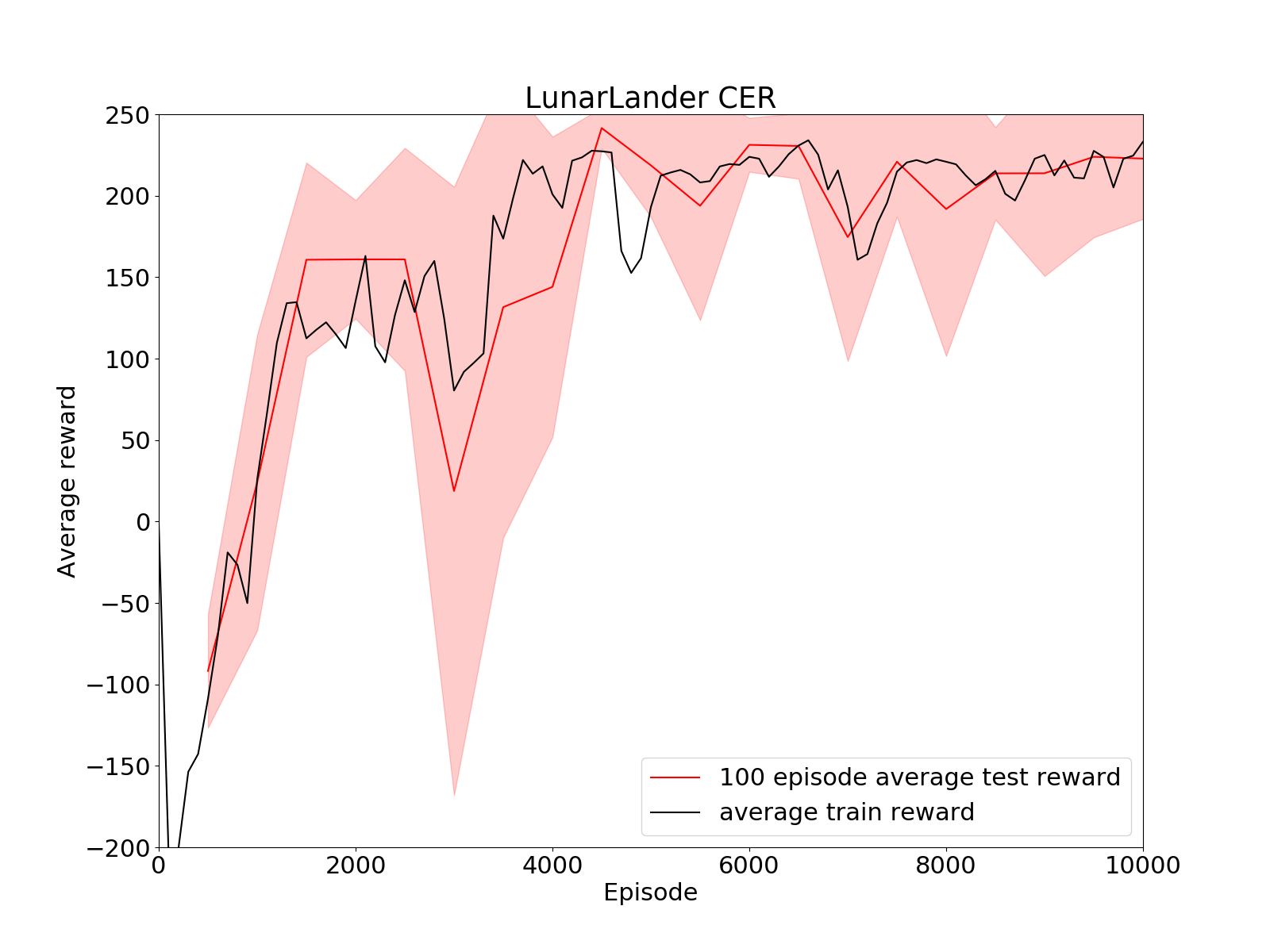}
\end{subfigure}
\begin{subfigure}[b]{0.45\textwidth}
\includegraphics[width=\textwidth]{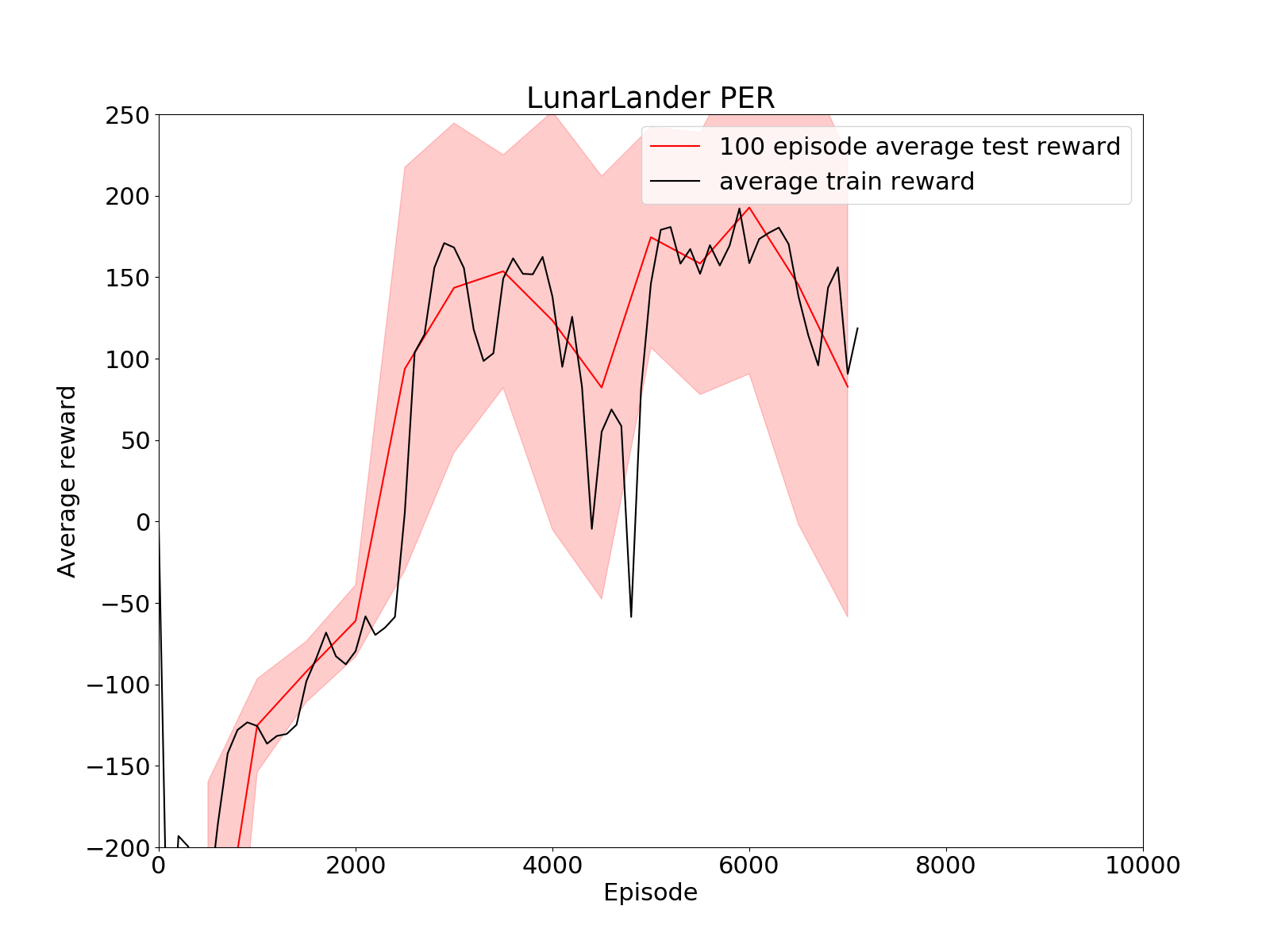}
\end{subfigure}%
\begin{subfigure}[b]{0.45\textwidth}
\includegraphics[width=\textwidth]{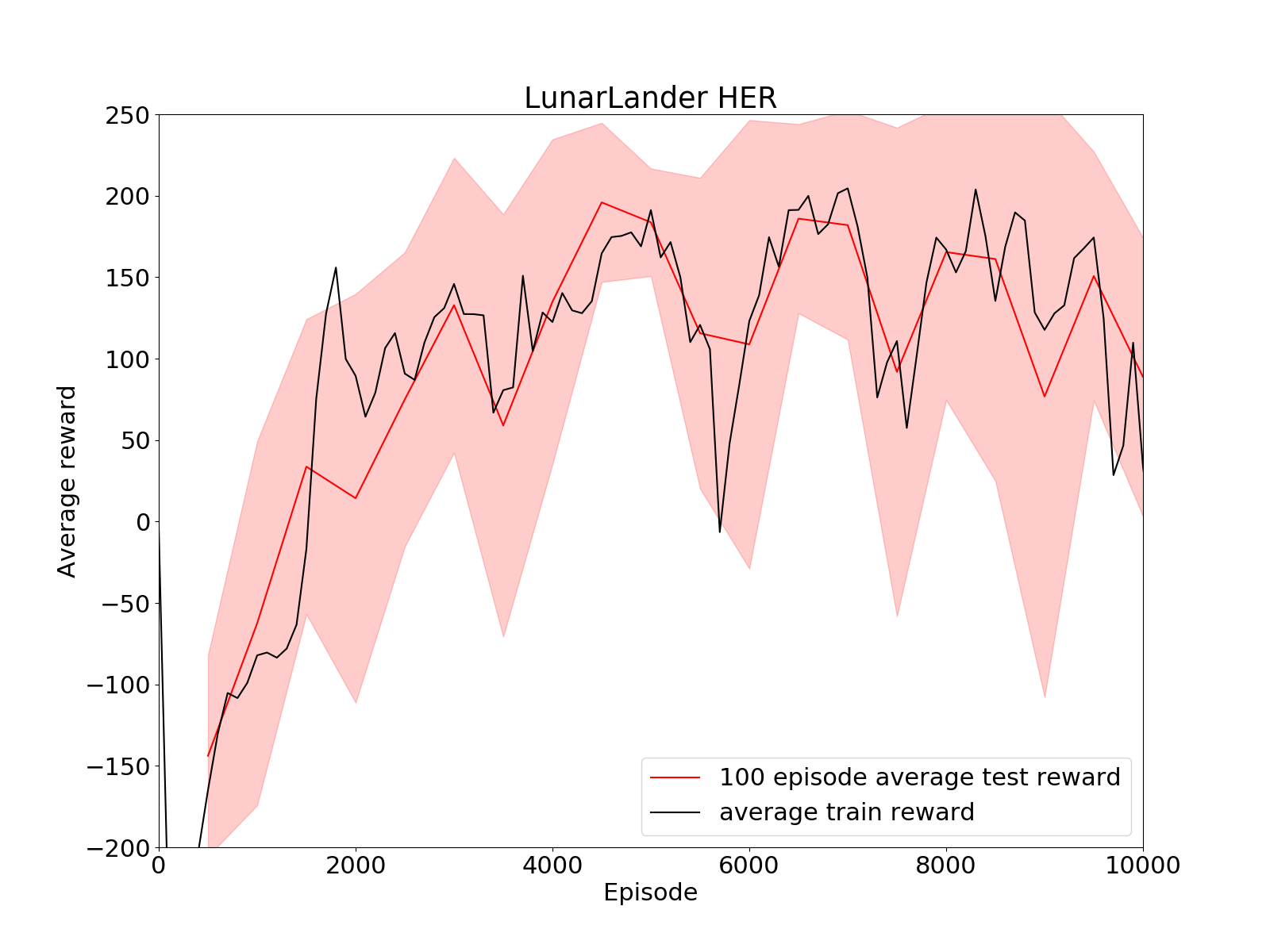}
\end{subfigure}
\end{figure}
\begin{figure}[htb]\ContinuedFloat
\begin{subfigure}[b]{0.45\textwidth}
\includegraphics[width=\textwidth]{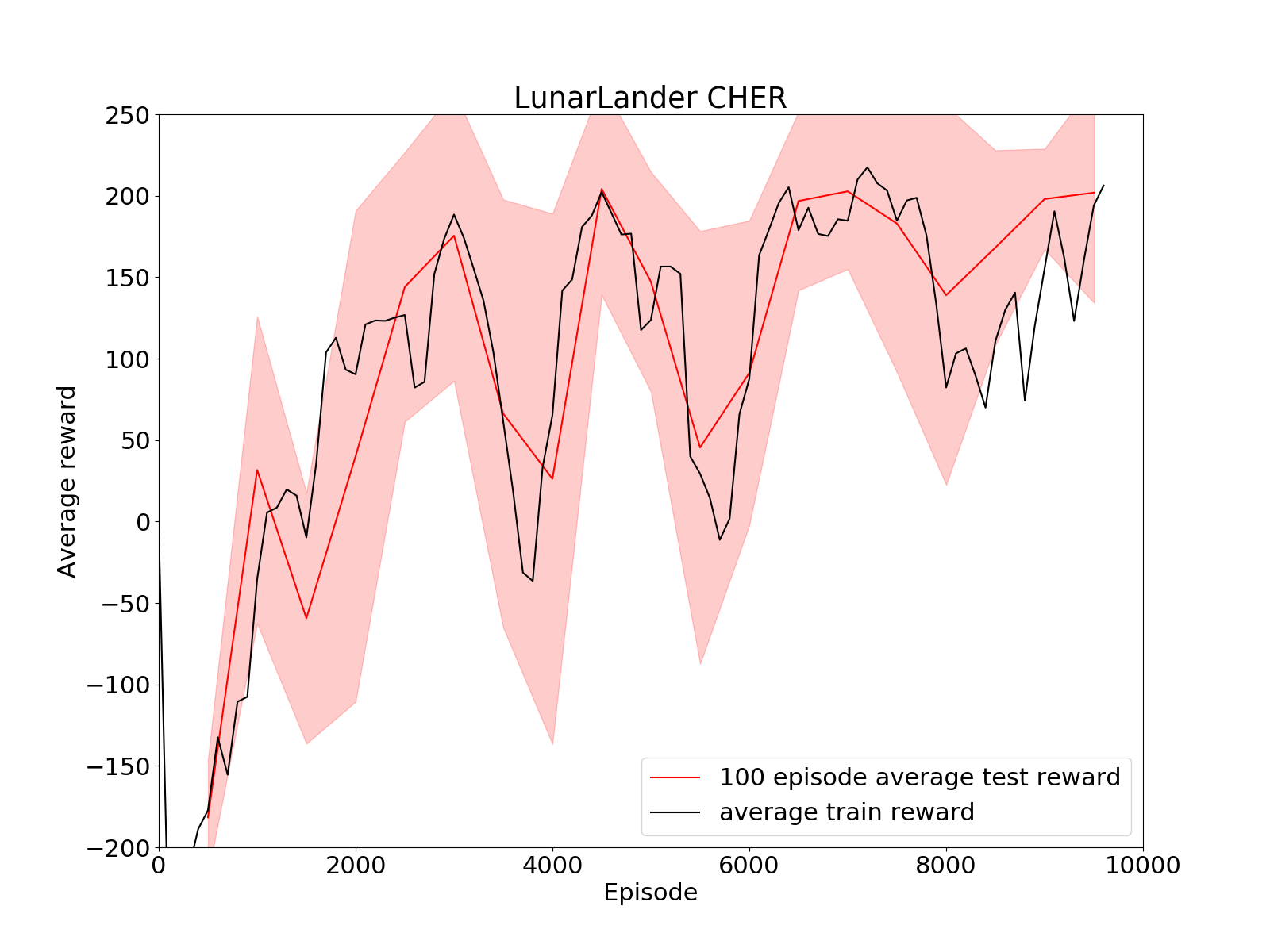}
\end{subfigure}%
\begin{subfigure}[b]{0.45\textwidth}
\includegraphics[width=\textwidth]{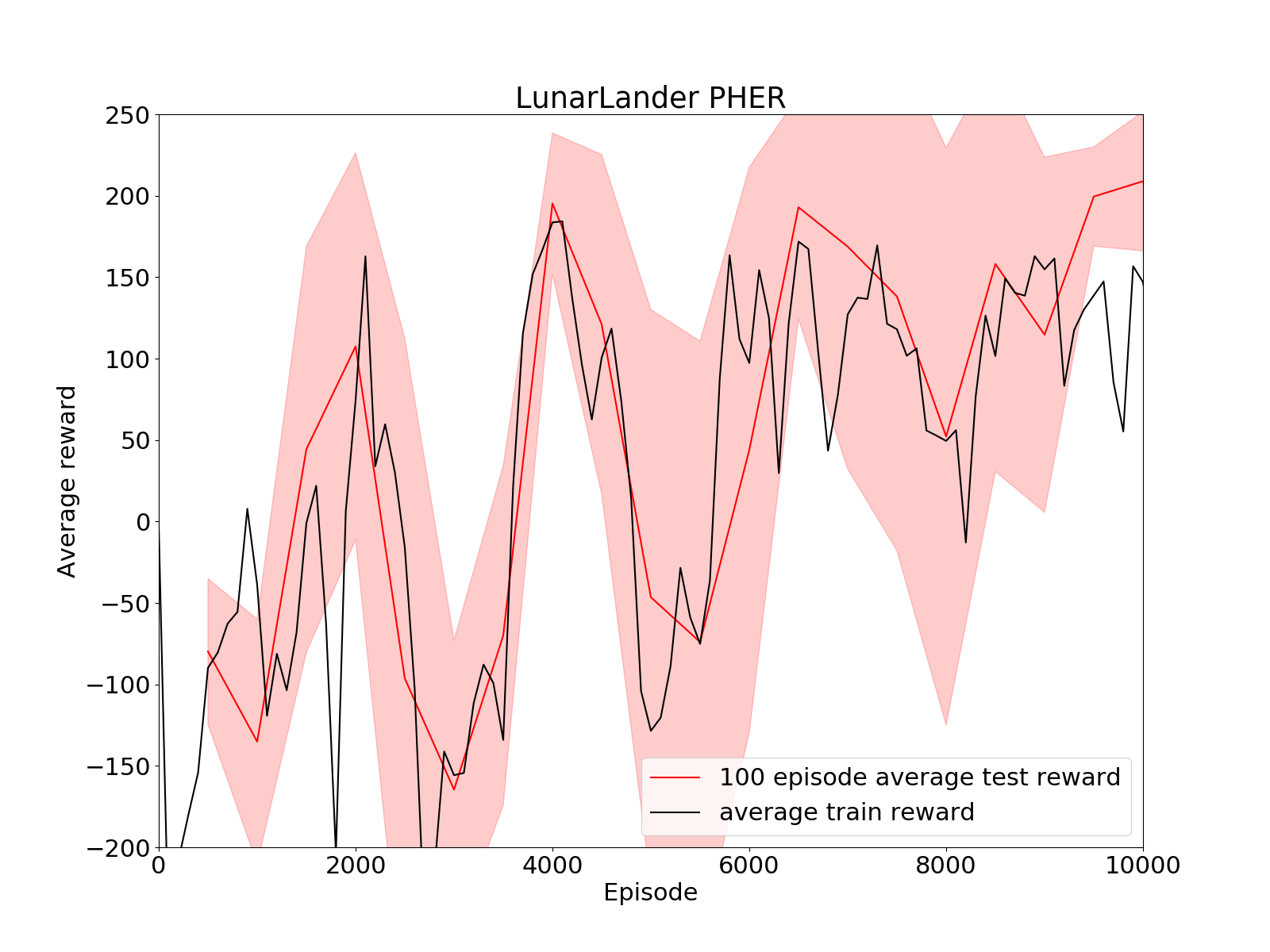}
\end{subfigure}
\begin{subfigure}[b]{0.45\textwidth}
\includegraphics[width=\textwidth]{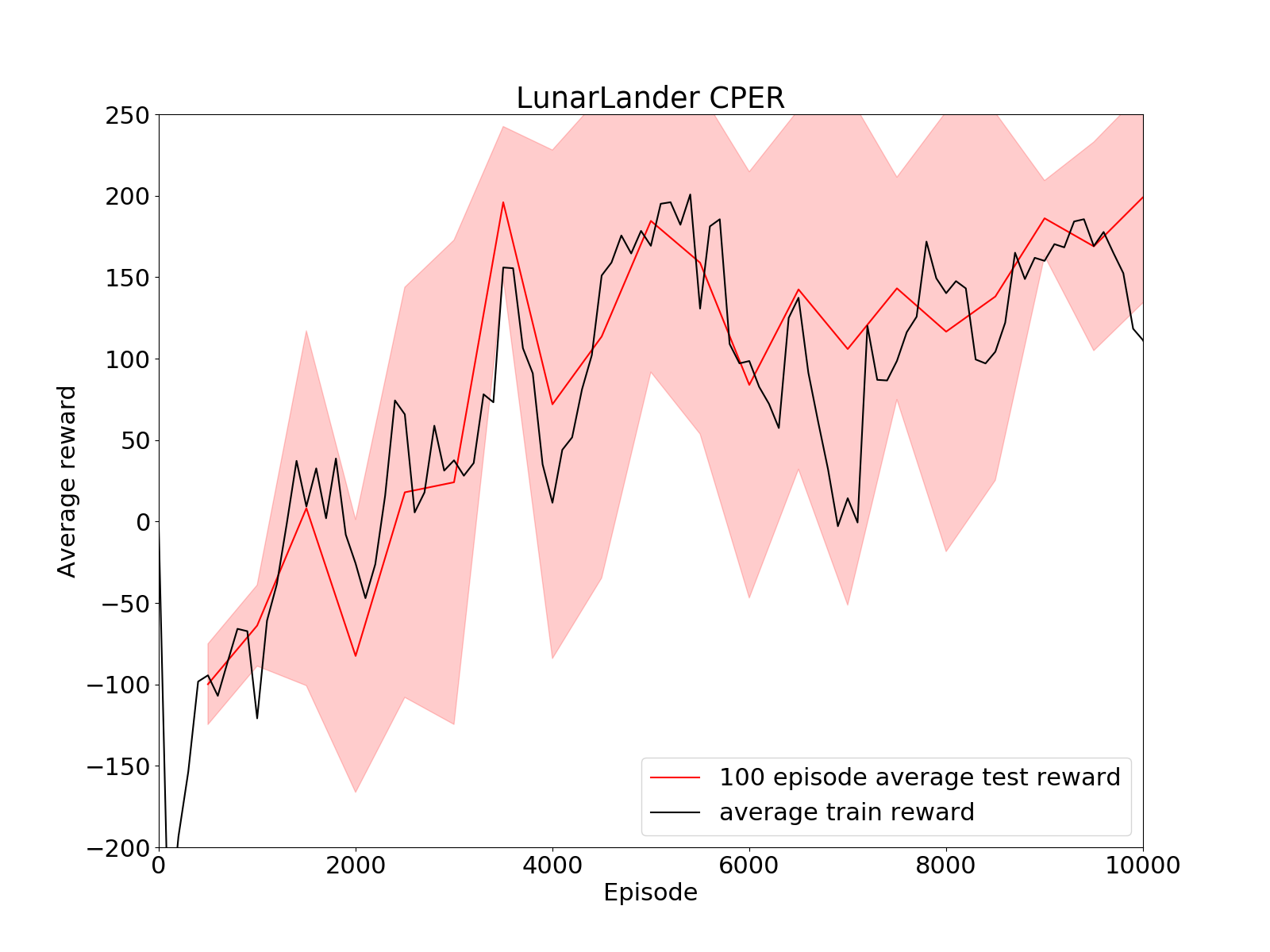}
\end{subfigure}%
\begin{subfigure}[b]{0.45\textwidth}
\includegraphics[width=\textwidth]{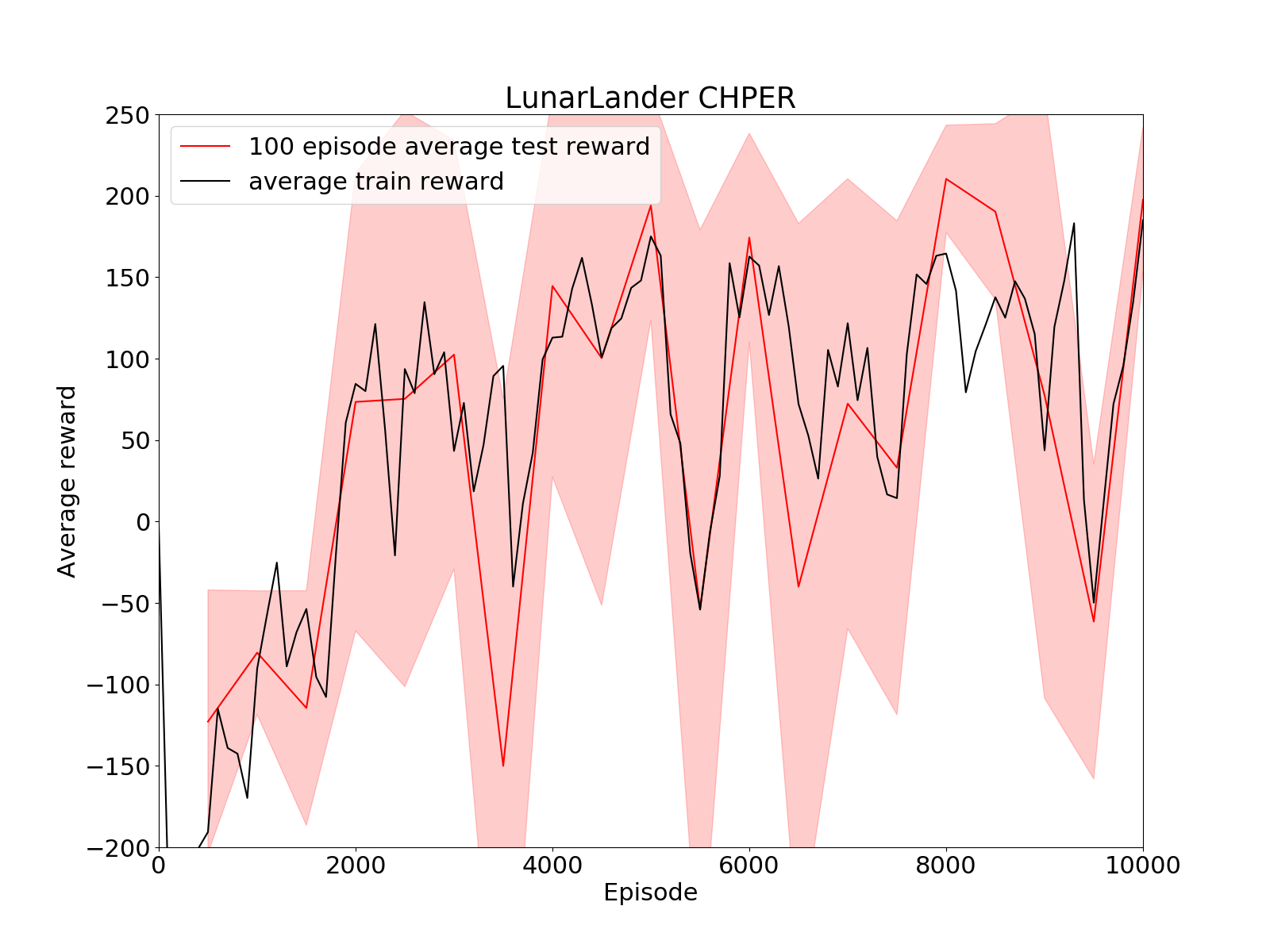}
\end{subfigure}
\caption{Training/test average reward over time of different experience replay strategies for the \texttt{LunarLander-v2} task. The red background is the standard deviation of the test reward.}
\end{figure}
\newpage
\begin{figure}[H]
\begin{subfigure}[b]{0.45\textwidth}
\includegraphics[width=\textwidth]{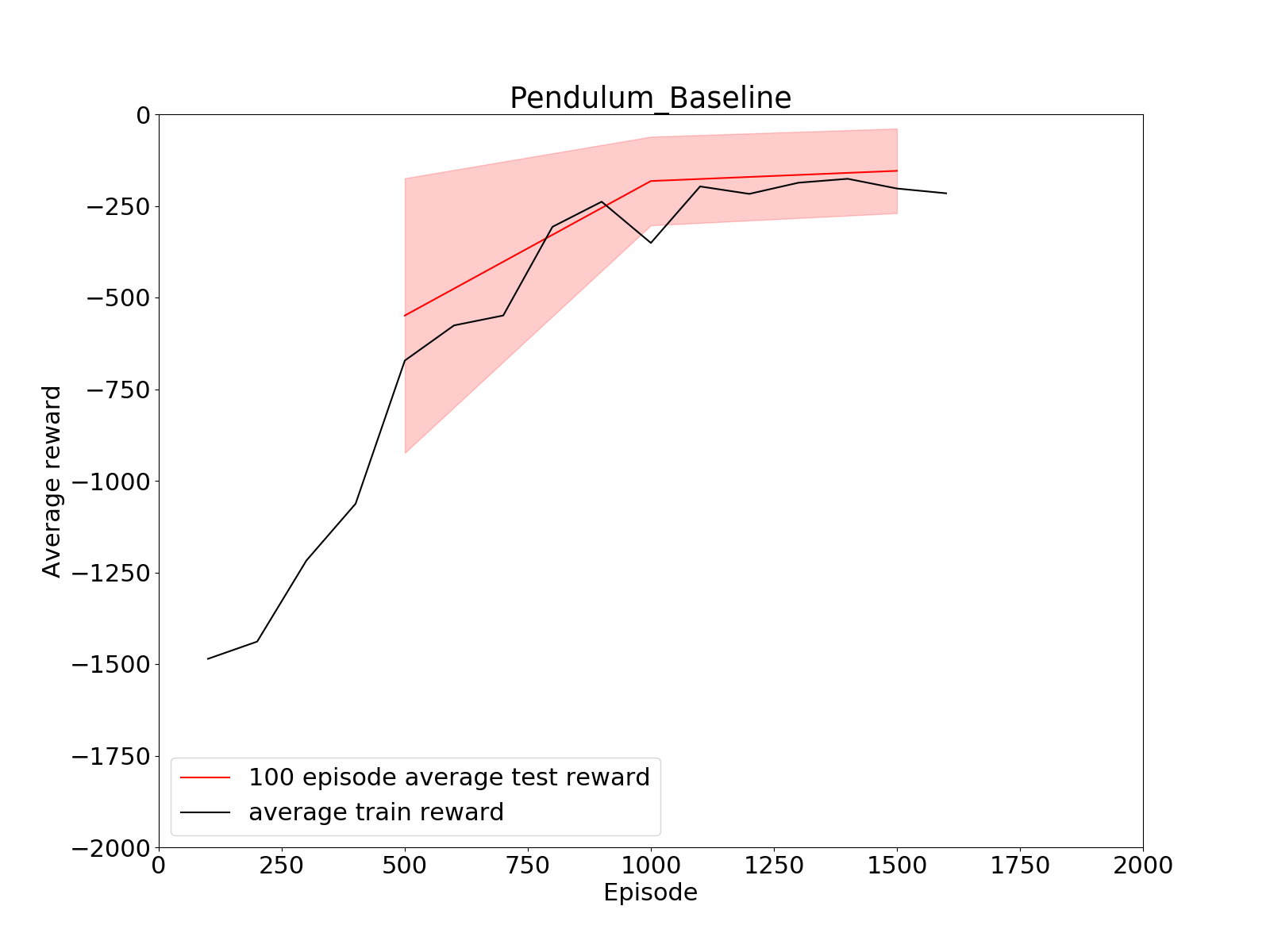}
\end{subfigure}%
\begin{subfigure}[b]{0.45\textwidth}
\includegraphics[width=\textwidth]{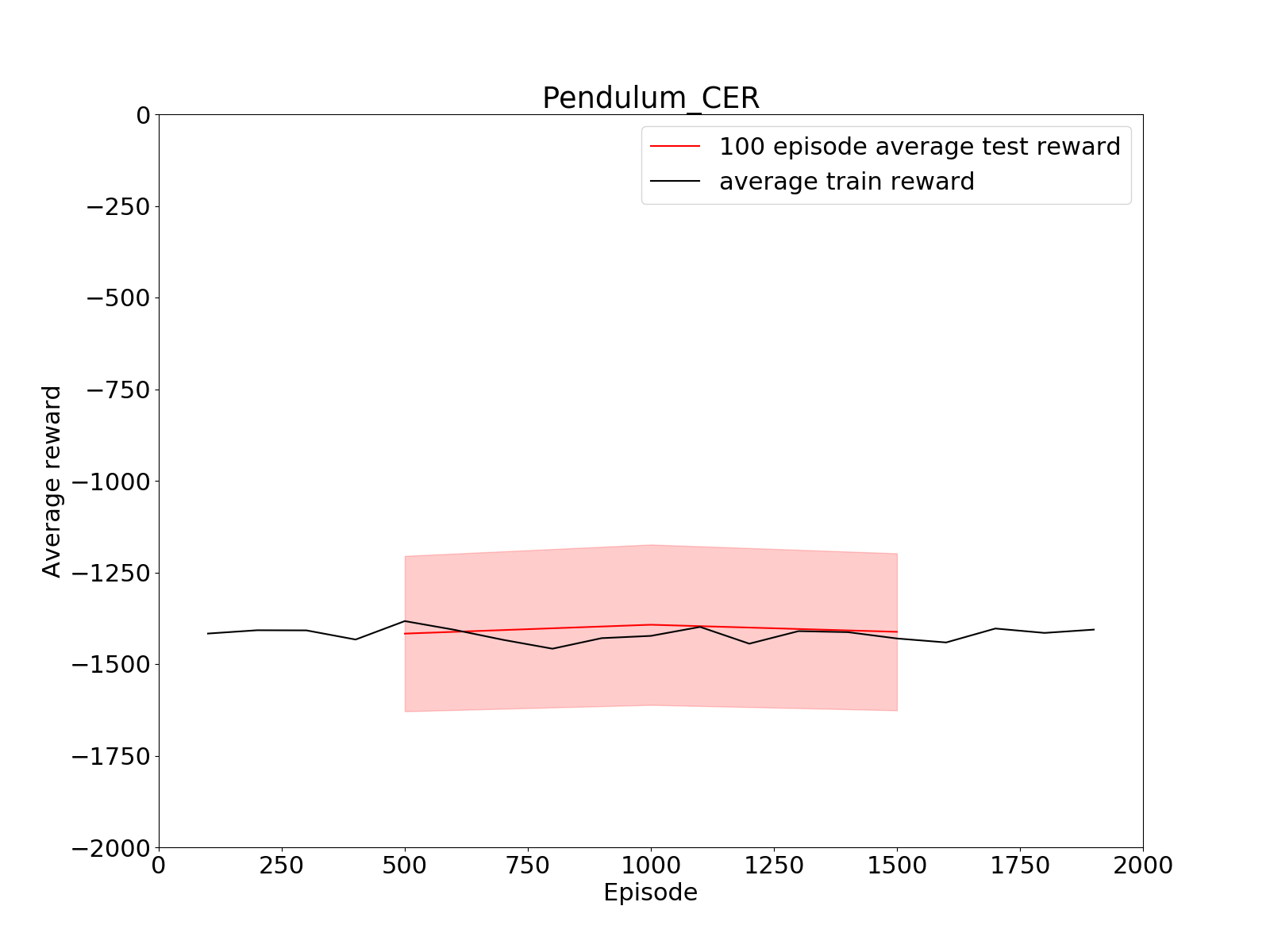}
\end{subfigure}
\begin{subfigure}[b]{0.45\textwidth}
\includegraphics[width=\textwidth]{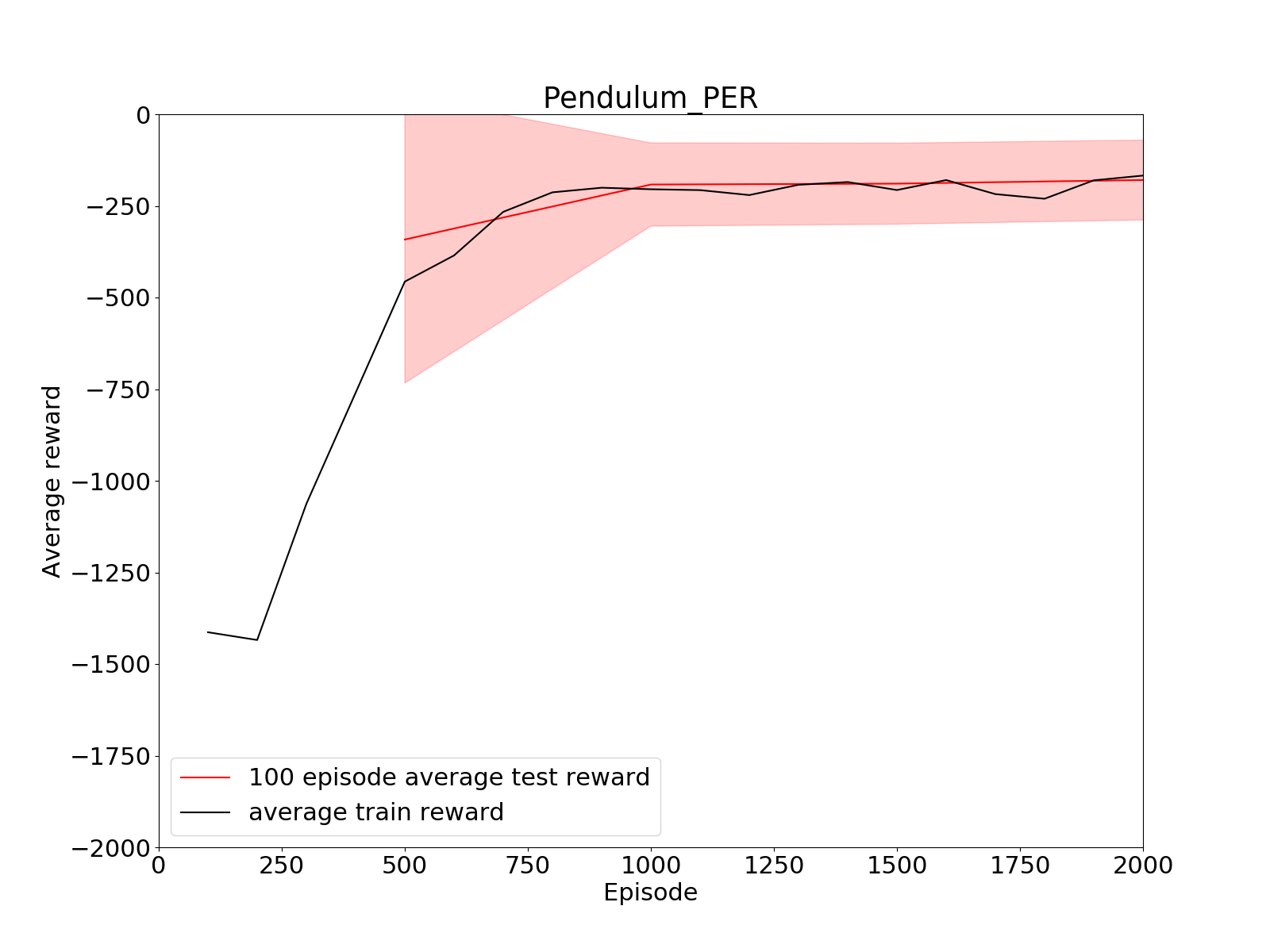}
\end{subfigure}%
\begin{subfigure}[b]{0.45\textwidth}
\includegraphics[width=\textwidth]{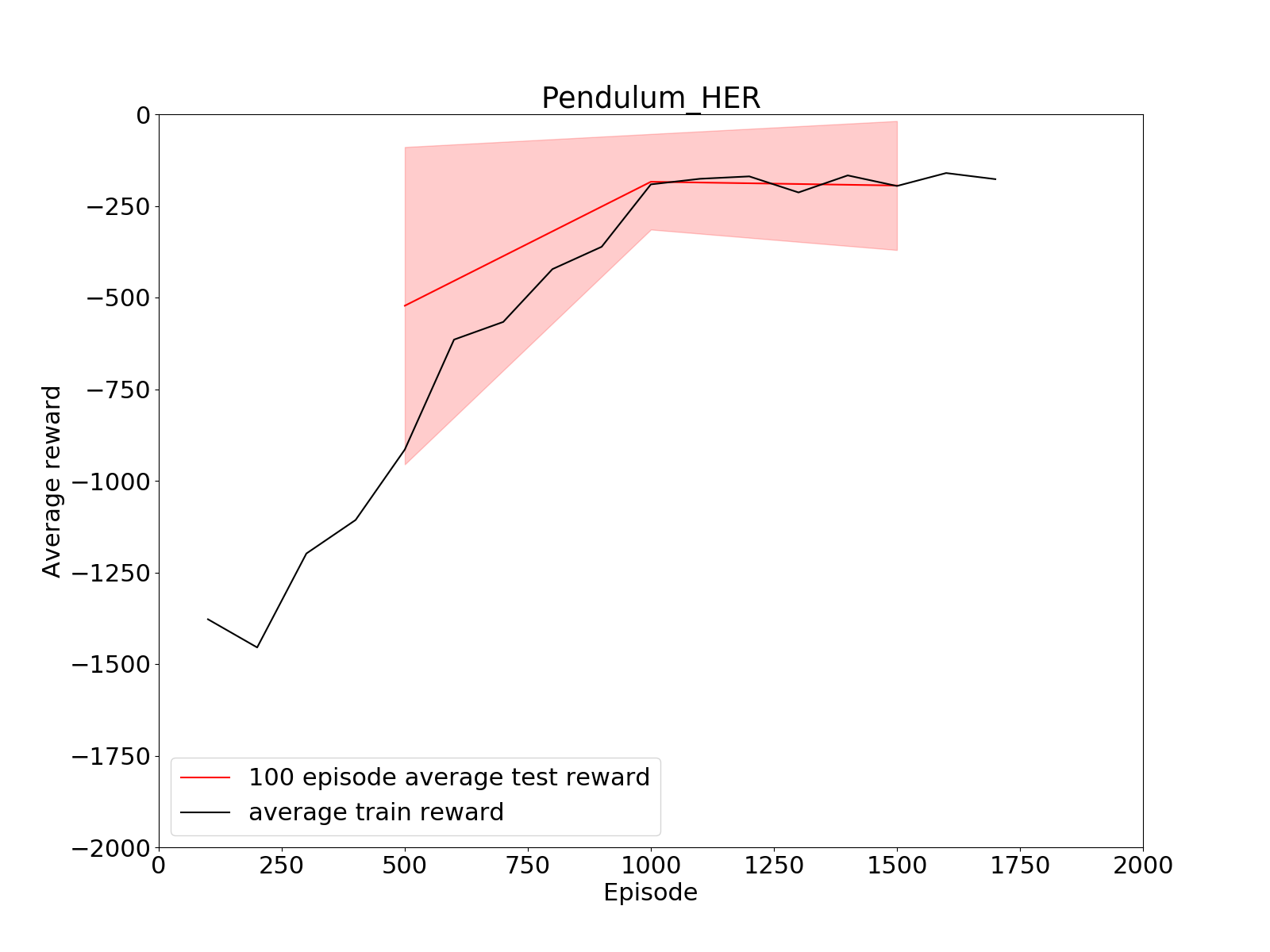}
\end{subfigure}
\end{figure}
\begin{figure}[htb]\ContinuedFloat
\begin{subfigure}[b]{0.45\textwidth}
\includegraphics[width=\textwidth]{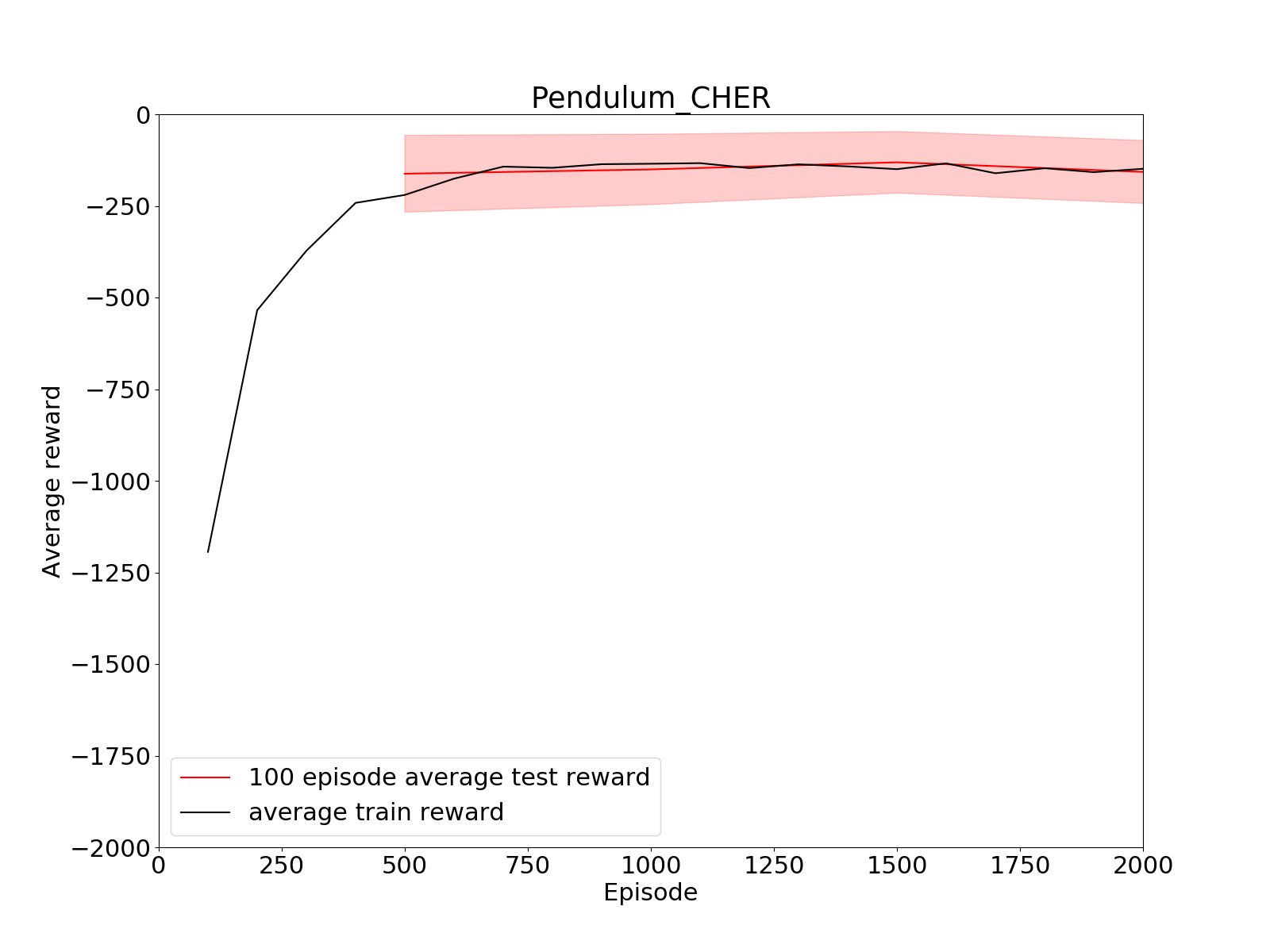}
\end{subfigure}%
\begin{subfigure}[b]{0.45\textwidth}
\includegraphics[width=\textwidth]{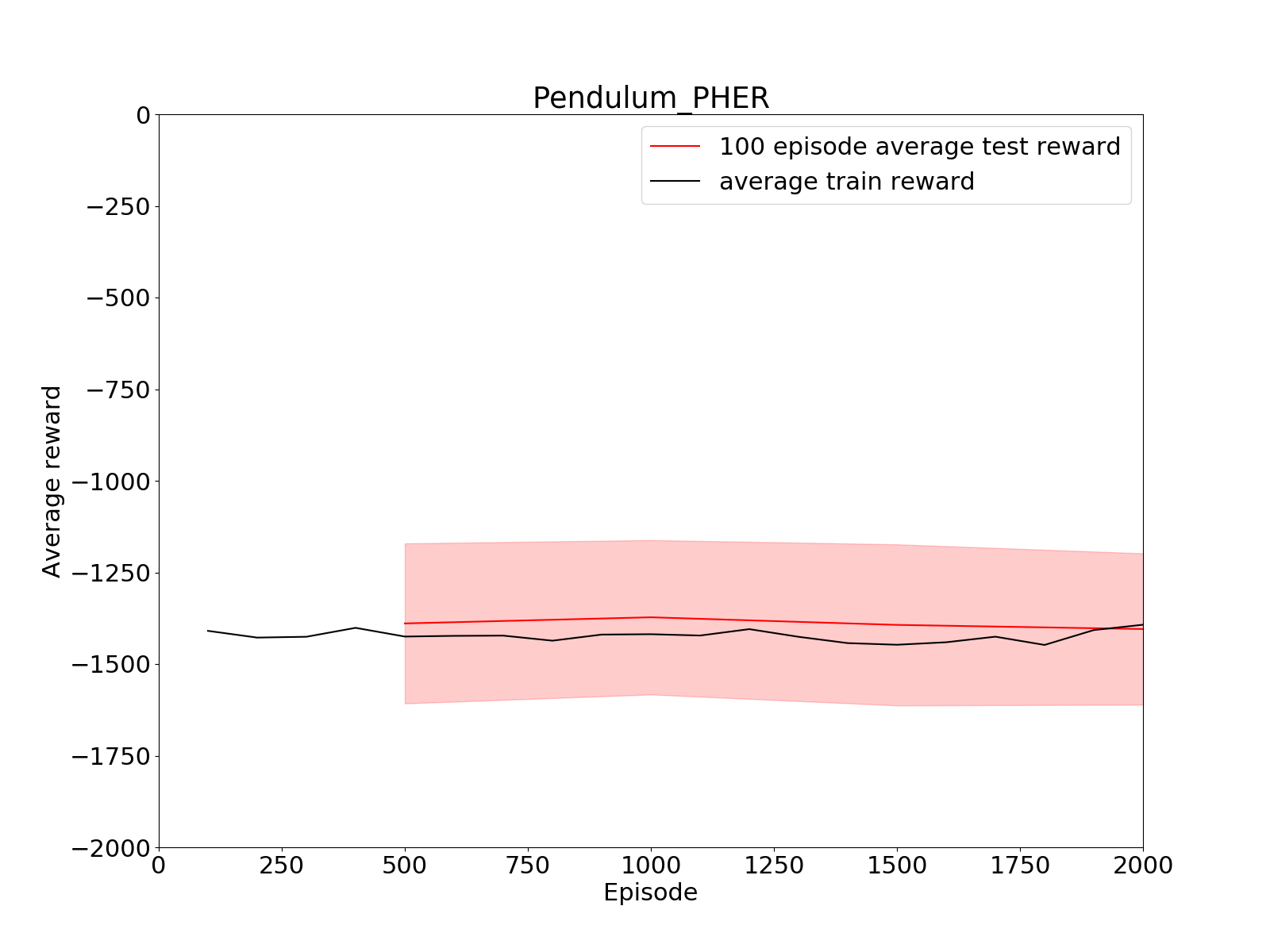}
\end{subfigure}
\begin{subfigure}[b]{0.45\textwidth}
\includegraphics[width=\textwidth]{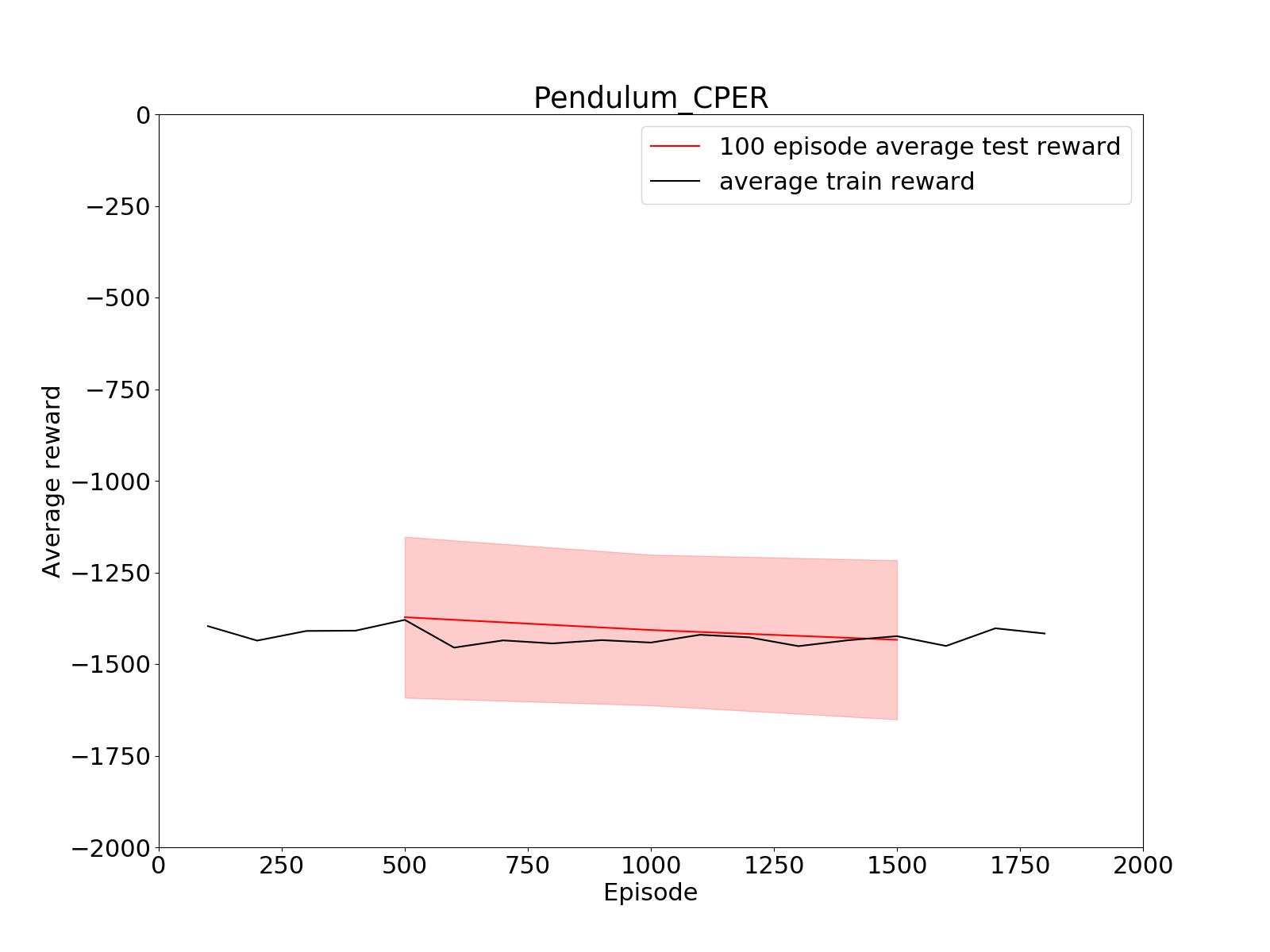}
\end{subfigure}%
\begin{subfigure}[b]{0.45\textwidth}
\includegraphics[width=\textwidth]{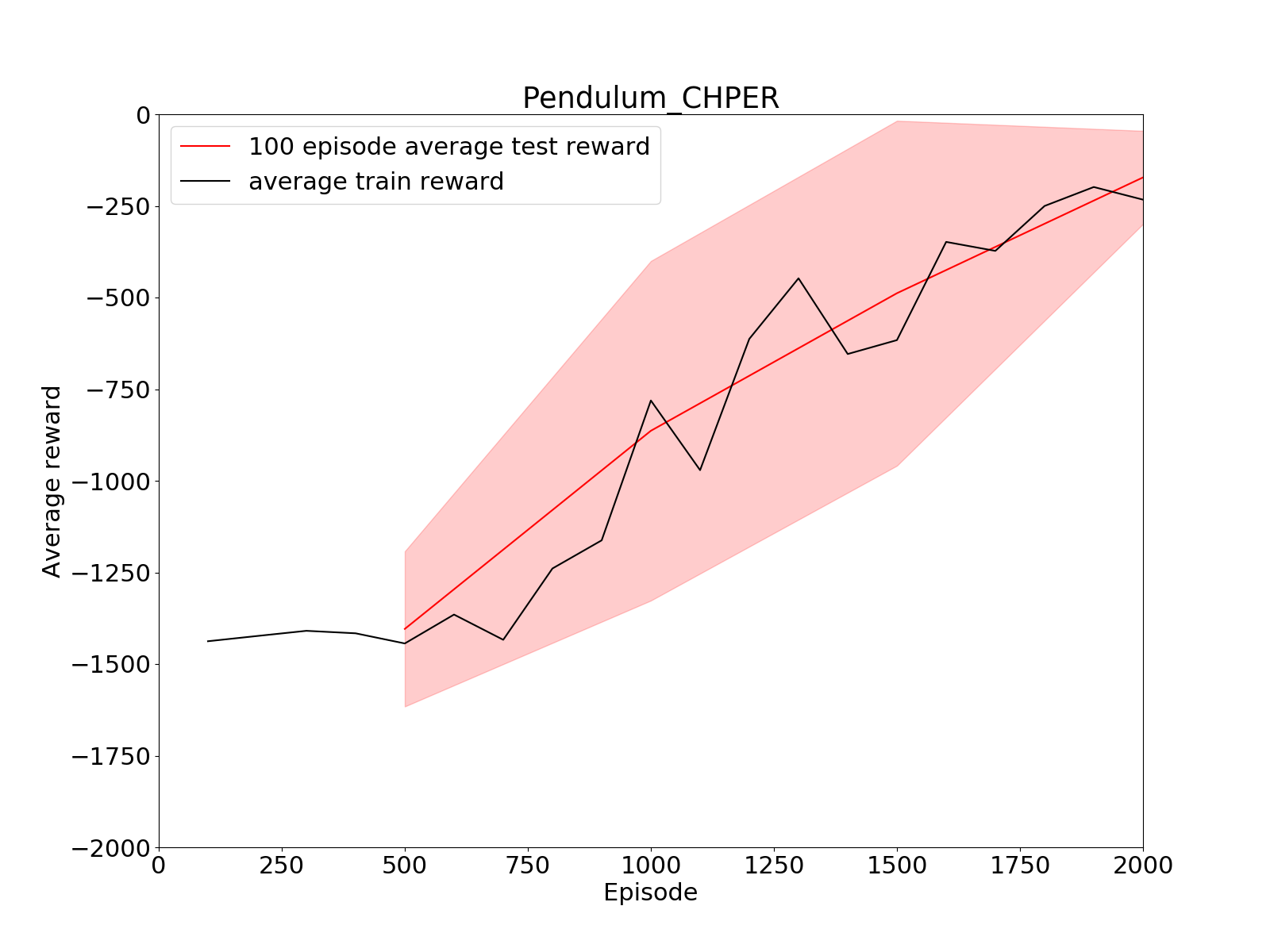}
\end{subfigure}
\caption{Training/test average reward over time of different experience replay strategies for the \texttt{Pendulum-v2} task. The red background is the standard deviation of the test reward.}
\end{figure}




\section{Discussion}
\subsection{CartPole-V0}
CartPole-v0 is a task of balancing a pole on top of the cart. The cart has access to its position and velocity as state, and can only go left or right for each action. The task is over when the pole falls over, the cart goes out of the boundaries, or 200 time steps are reached, with each step returning 1 reward. 

Prioritized and combined replay both ended up being detrimental to CartPole-v0 over the baseline. It may be that CartPole-v0 is an easily solved task, which results in additional techniques only perturbing the training, but show no additional benefit.

\subsection{MountainCar-V0}
MountainCar-V0 is a one dimensional track between two mountains. The goal is to drive up the mountain to the right. The agent receive a -1 reward for every time step it does not reach the top. The episode terminates when it reaches the top and receives 0 reward. The objective is for the agent to learn to drive back and forth to build momentum that will be enough to push the car up the hill. The observation consists of the car's position and velocity and the action consists of pushing left, pushing right and no push. For hindsight replay, the code passes in the position of the car as the modified goal. The main challenge of mountain car is that the rewards are very sparse since the agent only receive a reward if it reaches the top. 

We found that our MountainCar results were not very good, and there is not much of a differentiating factor among the different ER techniques tried, and so we omit the results for brevity.

\subsection{LunarLander-v2}
LunarLander-v2 is a two dimensional environment featuring a landing pad at (x,y)=(0,0). The goal is to move from the top of the screen to the landing pad without crashing. The agent receives 100-140 points for landing. If it moves away from the landing pad, it loses rewards. If it crashes it will receive -100 and if it lands successfully it receives 100.  Each leg ground contact is 10. Firing main engine is -0.3 points each frame. The observation consists of f the x and y coordinates, the x and y velocities, angle, angular velocity, and ground contact information of the lander and the action consists of do nothing, fire left orientation engine, fire main engine, fire right orientation engine.

An important distinguishing feature of the LunarLander-v2 environment and the other environments tested in this paper is that it is the only one that has dense rewards. The reward each step is calculated based on the distance of the lander way from its landing spot. Consequently, this environment may not suffer from the issues that hindsight and prioritized experience replay are best at tackling: sparse rewards and off policy sampling that does not assist the agent. We can observe that the results for techniques involving hindsight and prioritized experience replay do worse than baseline on this task, and this may be due to the non-universality of these techniques. In contrast, combined experience replay performed close to baseline, but still worse than the baseline. This perhaps is on account of the combined strategy being only slightly deviant from the baseline sampling strategy.

\subsection{Pendulum-v0}
Pendulum-v0 is a two dimensional environment featuring a frictionless pendulum. The goal is to keep the pendulum standing. The precise equation for reward is $-(\theta^2 + 0.1*\theta_{dt}^2 + 0.001*action^2).$ The observation consists of $\cos(\theta), \sin(\theta), $ and $\theta_{dt}.$ The action consists of the joint effort, which ranges between -2.0 and 2.0. For hindsight replay, we pass in the angle of the pendulum as the goal with the original goal being achieving a vertical position.  

Unlike all the other environments, Pendulum-v0 has continuous action space. We found that our DDPG algorithm did very poorly on this task. Many times the agent never reached the goal. We were supposed to train pendulum with baseline, P, H, CH, and CHP. Among them, CHER converged the fastest, at 500 episodes. This demonstrates that hindsight replay has great potential in continuous action environment. Further hyperparameter tuning could improve the stability of the results. 
\section{Conclusion}

Through this endeavor, we found that the field of reinforcement learning is highly unstable, and testing often requires multiple trials.  
Unfortunately this means that there was a large amount of variance in our results and so some experience replay methods which worked well at times would work terribly at other times. 

Our DDPG algorithm in particular did not perform well, and so we were unable to get substantial results on continuous action environments due to the poor training of the model. 

However, in conclusion, we believe that there is a lot of promise in combining the various experience replay techniques proposed in recent years. Hindsight experience replay in particular makes a lot of sense as a standalone technique, since it provides a lot more information to the agent in that the agent can still learn even if it ends in a different final goal state. Combining this with prioritized experience replay should only serve to improve the convergence of the agent by choosing more informative updates. And using the idea of always learning from the most recent experience a la combined experience replay should also add a slight edge. We believe that these techniques should be further explored and exploited in the future.


\printbibliography 

\end{document}